\pdfoutput=1
 \documentclass[10pt,twocolumn,letterpaper]{article}

\usepackage{cvpr}
\usepackage{times}
\usepackage{epsfig}
\usepackage{graphicx}
\usepackage{amsmath}
\usepackage{amssymb}
\usepackage{subcaption}
\usepackage{xcolor}
\usepackage[font=small,labelfont=bf,tableposition=top]{caption}
\usepackage{booktabs}
\usepackage{multirow}
\usepackage{algorithm,algorithmic}
\usepackage[numbers,sort]{natbib} 
\usepackage{authblk}

\DeclareMathOperator*{\argmin}{arg\,min}

\usepackage[pagebackref=true,breaklinks=true,letterpaper=true,colorlinks,bookmarks=false]{hyperref}

\cvprfinalcopy 

\newcommand{\methodlong}{temporal cycle-consistency }
\newcommand{\blocka}[2]{
  \(\left[
      \begin{array}{c}
        \text{3$\times$3, #1}\\[-.1em]
        \text{3$\times$3, #1}
      \end{array}
    \right]\)$\times$#2
}
\newcommand{\blockb}[3]{
  \(\left[
      \begin{array}{c}
        \text{1$\times$1, #1}\\[-.1em]
        \text{3$\times$3, #1}\\[-.1em]
        \text{1$\times$1, #2}
      \end{array}
    \right]\)$\times$#3
}

\newcommand{\blockc}[3]{
  \(\left[
      \begin{array}{c}
        \text{3$\times$3$\times$3, #1}\\[-.1em]
        \text{3$\times$3$\times$3, #1}
      \end{array}
    \right]\)$\times$#2
}

\newcommand{\blockd}[2]{
  \(\left[
      \begin{array}{c}
        \text{#1}\\[-.1em]
        \text{#1}
      \end{array}
    \right]\)$\times$#2
}

\makeatletter
\renewcommand\AB@affilsepx{  \protect\Affilfont}
\makeatother

\begin{document}

\title{Temporal Cycle-Consistency Learning \vspace{-1em}}

\author[ 1]{Debidatta Dwibedi}
\author[ 2]{Yusuf Aytar}
\author[ 1]{Jonathan Tompson}
\author[ 1]{Pierre Sermanet}
\author[ 2]{Andrew Zisserman}
\affil[ 1 ]{Google Brain}
\affil[ 2 ]{DeepMind\protect\\\tt\small \{debidatta, yusufaytar, tompson, sermanet, zisserman\}@google.com}

\twocolumn[{%
\renewcommand\twocolumn[1][]{#1}%
\maketitle
\vspace{-3.5em}
\begin{center}
\includegraphics[width=\textwidth]{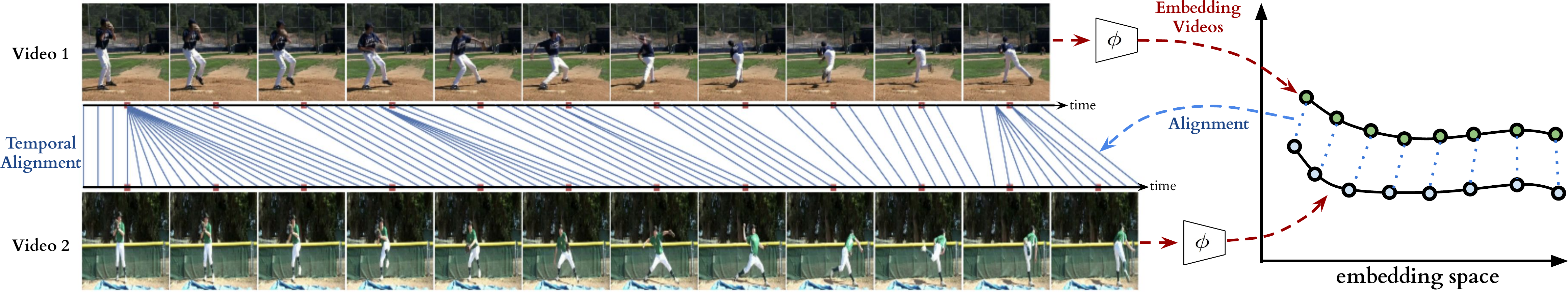}
\captionof{figure}{We present a self-supervised representation learning technique called temporal cycle consistency (TCC) learning. It is inspired by the temporal video alignment problem, which refers to the task of finding correspondences across multiple videos despite many factors of variation. The learned representations are useful for fine-grained temporal understanding in videos. Additionally, we can now align multiple videos by simply finding nearest-neighbor frames in the embedding space. 
}
\label{fig:teaser}
\end{center}%
}]

\begin{abstract}

We introduce a self-supervised representation learning method based on the task of
temporal alignment between videos. The method trains a network using \methodlong (TCC), a differentiable cycle-consistency loss that can be used to find correspondences across time in multiple videos. The resulting per-frame embeddings can be used to align videos by simply matching frames using nearest-neighbors in the learned embedding space.

To evaluate the power of the embeddings, we densely label the \textit{Pouring} and \textit{Penn Action} video datasets for action phases. We show that (i) the learned embeddings enable few-shot classification of these action phases, significantly reducing the supervised training requirements; and (ii) TCC is complementary to other methods of self-supervised learning in videos, such as Shuffle and Learn and Time-Contrastive Networks. The embeddings are also used for a number of applications based on alignment (dense temporal correspondence) between video pairs, including transfer of metadata of synchronized modalities between videos (sounds, temporal semantic labels), synchronized playback of multiple videos, and anomaly detection. Project webpage: \url{https://sites.google.com/view/temporal-cycle-consistency}.
\end{abstract}


\section{Introduction}
The world presents us with abundant examples of sequential processes. A plant growing from a seedling to a tree, the daily routine of getting up, going to work and coming back home, or a person pouring themselves a glass of water -- are all examples of events that happen in a particular order. Videos capturing such processes not only contain information about the causal nature of these events, but also provide us with a valuable signal -- the possibility of temporal \textit{correspondences} lurking across multiple instances of the same process. 
For example, during pouring, one could be reaching for a teapot, a bottle of wine, or a glass of water to pour from. Key moments such as the first touch to the container or the container being lifted from the ground are common to all pouring sequences. These correspondences, which exist in spite of many varying factors like visual changes in viewpoint, scale, container style, the speed of the event, etc., could serve as the link between raw video sequences and high-level temporal abstractions (e.g.\ phases of actions). In this work we present evidence that suggests the very act of \textit{looking for correspondences} in sequential data enables the learning of rich and useful representations, particularly suited for fine-grained temporal understanding of videos.

Temporal reasoning in videos, understanding multiple stages of a process and causal relations between them, is a relatively less studied problem compared to recognizing action categories \cite{carreira2017quo,soomro2012ucf101}. Learning representations that can differentiate between states of objects as an action proceeds is critical for perceiving and acting in the world. It would be desirable for a robot tasked with learning to pour drinks to understand each intermediate state of the world as it proceeds with performing the task. Although videos are a rich source of sequential data essential to understanding such state changes, their true potential remains largely untapped. One hindrance in the fine-grained temporal understanding of videos can be an excessive dependence on pure supervised learning methods that require per-frame annotations. It is not only difficult to get every frame labeled in a video because of the manual effort involved, but also it is not entirely clear what are the exhaustive set of labels that need to be collected for fine-grained understanding of videos. Alternatively, we explore self-supervised learning of correspondences between videos across time. We show that the emerging features have strong temporal reasoning capacity, which is demonstrated through tasks such as action phase classification and tracking the progress of an action.   

When frame-by-frame alignment (i.e.\ supervision) is available, learning correspondences reduces to learning a common embedding space from pairs of aligned frames (e.g.\ CCA~\cite{anderson1958introduction,andrew2013deep} and ranking loss~\cite{Sermanet2017TCN}). However, for most of the real world sequences such frame-by-frame alignment does not exist naturally. One option would be to artificially obtain aligned sequences by recording the same event through multiple cameras~\cite{Sermanet2017TCN,sigurdsson2018actor,revaud2013event}. Such data collection methods might find it difficult to capture all the variations present naturally in videos in the wild. On the other hand, our self-supervised objective does not need explicit correspondences to align different sequences. It can align significant variations within an action category (e.g.\ pouring liquids, or baseball pitch). Interestingly, the embeddings that emerge from learning the alignment prove to be useful for fine-grained temporal understanding of videos. More specifically, we learn an embedding space that maximizes one-to-one mappings (i.e. cycle-consistent points) across pairs of video sequences within an action category. In order to do that, we introduce two differentiable versions of cycle consistency computation which can be optimized by conventional gradient-based optimization methods. Further details of the method will be explained in section \ref{sec:method}.

The main contribution of this paper is a new self-supervised training method, referred to as temporal cycle consistency (TCC) learning, that learns representations by aligning video sequences of the same action. We compare TCC representations against features from existing self-supervised video representation methods \cite{Sermanet2017TCN,misra2016shuffle} and supervised learning, for the tasks of action phase classification and continuous progress tracking of an action. Our approach provides significant performance boosts when there is a lack of labeled data. We also collect per-frame annotations of Penn Action~\cite{zhang2013actemes} and Pouring~\cite{Sermanet2017TCN} datasets that we will release publicly to facilitate evaluation of fine-grained video understanding tasks.

\section{Related Work}

\noindent\textbf{Cycle consistency}. Validating good matches by cycling between two or more samples is a commonly used technique in computer vision. It has been applied successfully for tasks like co-segmentation~\cite{wang2014unsupervised,wang2013image}, structure from motion ~\cite{zach2010disambiguating,wilson2013network}, and image matching~\cite{zhou2015multi,zhou2016learning,zhou2015flowweb}.  
For instance, FlowWeb~\cite{zhou2015flowweb} optimizes globally-consistent
dense correspondences using the cycle consistent flow fields between all pairs of images in a collection, whereas Zhou et al.~\cite{zhou2015multi} approaches a similar task by formulating it as a low-rank matrix recovery problem and solves it through fast alternating minimization. These methods learn robust dense correspondences on top of fixed feature representations (e.g. SIFT, deep features, etc.) by enforcing cycle consistency and/or spatial constraints between the images. Our method differs from these approaches in that TCC is a self-supervised representation learning method which learns embedding spaces that are optimized to give good correspondences. Furthermore we address a temporal correspondence problem rather than a spatial one.
Zhou et al.~\cite{zhou2016learning} learn to align multiple images using the supervision from 3D guided cycle-consistency by leveraging the initial correspondences that are available between multiple renderings of a 3D model, whereas we don't assume any given correspondences. Another way of using cyclic relations is to directly learn bi-directional transformation functions between multiple spaces such as CycleGANs~\cite{zhu2017unpaired} for learning image transformations, and CyCADA~\cite{hoffman2017cycada} for domain adaptation. Unlike these approaches we don't have multiple domains, and we can't learn transformation functions between all pairs of sequences. Instead we learn a joint embedding space in which the Euclidean distance defines the mapping across the frames of multiple sequences.  Similar to us, Aytar et al.~\cite{aytar2018playing} applies cycle-consistency between temporal sequences, however they use it as a validation tool for hyper-parameter optimization of learned representations for the end goal of imitation learning. Unlike our approach, their cycle-consistency measure is non-differentiable and hence can't be directly used for representation learning. 

\noindent\textbf{Video alignment}. When we have synchronization information (e.g.\ multiple cameras recording the same event) then learning a mapping between multiple video sequences can be accomplished by using existing methods such as Canonical Correlation Analysis (CCA)~\cite{anderson1958introduction,andrew2013deep}, ranking~\cite{Sermanet2017TCN} or match-classification~\cite{arandjelovic2017look} objectives. For instance TCN~\cite{Sermanet2017TCN} and circulant temporal encoding~\cite{revaud2013event} align multiple views of the same event, whereas Sigurdsson et al.\cite{sigurdsson2018actor} learns to align first and third person videos. Although we have a similar objective, these methods are not suitable for our task as we cannot assume any given correspondences between different videos.

\noindent\textbf{Action localization and parsing}. As action recognition is quite popular in the computer vision community, many studies
\cite{wang2016temporal,sigurdsson2017asynchronous,zhao2017temporal,girdhar2017actionvlad,yeung2018every} explore efficient deep architectures for action recognition and localization in videos.  Past work has also explored parsing of fine-grained actions in videos \cite{pirsiavash2014parsing,lan2015action,lea2016segmental} while some others 
\cite{shechtman2007matching,del2015articulated, sener2015unsupervised,sener2018unsupervised} discover sub-activities without explicit supervision of temporal boundaries. \cite{heidarivincheh2018action} learns a supervised regression model with voting to predict the completion of an action, and \cite{Alayrac16unsupervised} discovers key events in an unsuperivsed manner using a weak association between videos and text instructions. 
However all these methods heavily rely on existing deep image \cite{he2016deep,simonyan2014very} or spatio-temporal~\cite{wang2013action} features, whereas we learn our representation from scratch using raw video sequences.    

\noindent\textbf{Soft nearest neighbours}. The differentiable or soft formulation for nearest-neighbors is a commonly known method \cite{goldberger2005neighbourhood}. This formulation has recently found application in metric learning for few-shot learning~\cite{snell2017prototypical, movshovitz2017no,rocco2018neighbourhood}. We also make use of soft nearest neighbor formulation as a component in our differentiable cycle-consistency computation.

\noindent\textbf{Self-supervised representations.} There has been significant progress in learning from images and videos without requiring class or temporal segmentation labels. Instead of labels, self-supervised learning methods use signals such as temporal order~\cite{misra2016shuffle,fernando2017self}, consistency across viewpoints and/or temporal neighbors~\cite{Sermanet2017TCN}, classifying arbitrary temporal segments~\cite{hyvarinen2016unsupervised}, temporal distance classification within or across modalities~\cite{aytar2018playing}, spatial permutation of patches~\cite{doersch2015unsupervised,anoop33deeppermnet}, visual similarity~\cite{sanakoyeu2018deep} or a combination of such signals~\cite{doersch2017multi}.
While most of these approaches optimize each sample independently, TCC jointly optimizes over two sequences at a time, potentially capturing more variations in the embedding space. Additionally, we show that TCC yields best results when combined with some of the unsupervised losses above.

\section{Cycle Consistent Representation Learning}
\label{sec:method}

\begin{figure}[t!]
\centering
\includegraphics[width=0.5\textwidth]{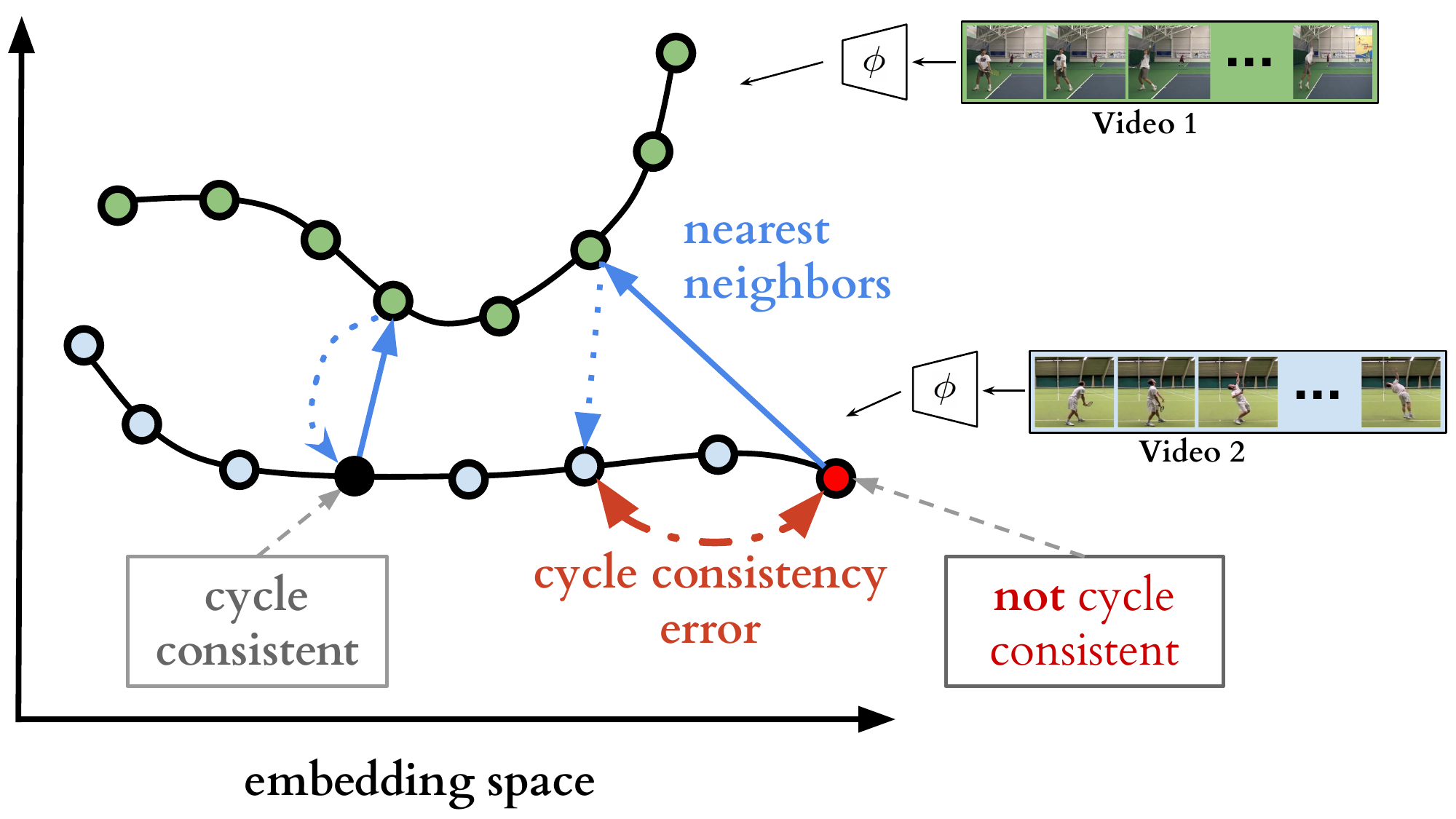}
\caption{{\bf Cycle-consistent representation learning.} We show two example video sequences encoded in an example embedding space. If we use nearest neighbors for matching, one point (shown in black) is \textit{cycling back to itself} while another one (shown in red) is not. Our target is to learn an embedding space where maximum number of points can cycle back to themselves. We achieve it by minimizing the cycle consistency error (shown in red dotted line) for each point in every pair of sequences.}
\label{fig:cycle}
\end{figure}

\begin{figure*}[!t]
\centering
\includegraphics[width=\textwidth]{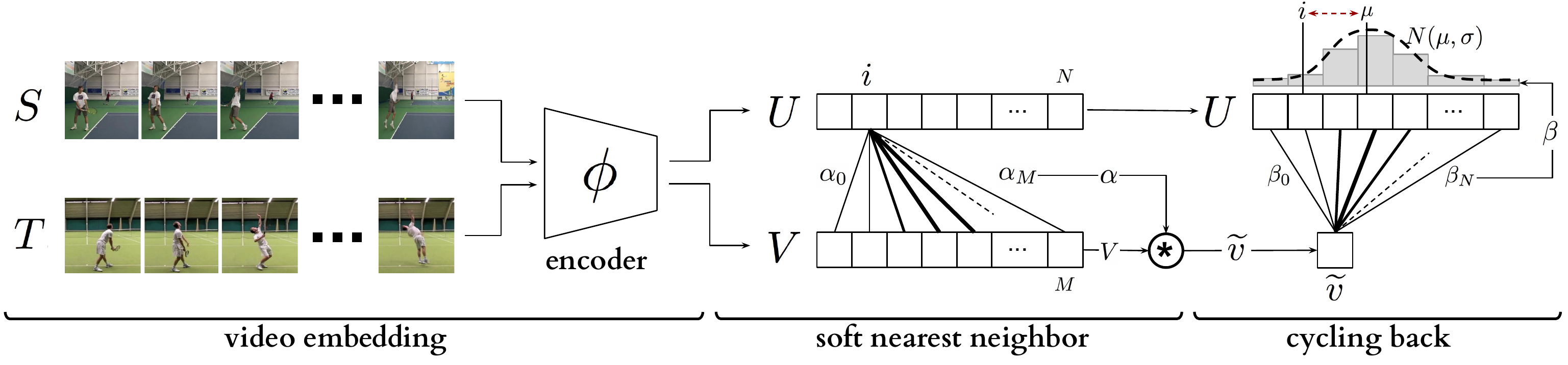}
\caption{{\bf Temporal cycle consistency}. The embedding sequences $U$ and $V$ are obtained by encoding video sequences $S$ and $T$ with the encoder network $\phi$, respectively. For the selected point $u_i$ in $U$, soft nearest neighbor computation and cycling back to $U$ again is demonstrated visually. Finally the normalized distance between the index $i$ and cycling back distribution $N(\mu,\sigma^2)$ (which is fitted to $\beta$) is minimized.}
\label{fig:soft_cycle_consistency}
\end{figure*}

The core contribution of this work is a self-supervised approach to learn an embedding space where two similar video sequences can be aligned temporally. More specifically, we intend to maximize the number of points that can be mapped one-to-one between two sequences by using the minimum distance in the learned embedding space. We can achieve such an objective by maximizing the number of cycle-consistent frames between two sequences (see Figure~\ref{fig:cycle}). However, cycle-consistency computation is typically not a differentiable procedure. In order to facilitate learning such an embedding space using back-propagation, we introduce two differentiable versions of the \textit{cycle-consistency loss}, which we describe in detail below.

Given any frame $s_i$ in a sequence $S=\{s_1,s_2,...,s_N\}$, the embedding is computed as $u_i = \phi(s_i;\theta)$, where $\phi$ is the neural network encoder parameterized by $\theta$. For the following sections, assume we are given two video sequences $S$ and $T$, with lengths $N$ and $M$, respectively. Their embeddings are computed as $U=\{u_1,u_2,...,u_N\}$ and $V=\{v_1,v_2,...,v_M\}$ such that $u_i = \phi(s_i;\theta)$ and $v_i = \phi(t_i;\theta)$.

\subsection{Cycle-consistency}
\label{sec:cycle_consistency}
In order to check if a point $u_i \in U$ is cycle consistent, we first determine its nearest neighbor, $v_j = \argmin_{v \in V} ||u_i-v||$. We then repeat the process to find the nearest neighbor of $v_j$ in $U$, i.e. $u_k = \argmin_{u \in U} ||v_j-u||$. The point $u_i$ is \emph{cycle-consistent} if and only if $i=k$, in other words if the point $u_i$ cycles back to itself.
Figure \ref{fig:cycle} provides positive and negative examples of cycle consistent points in an embedding space.
We can learn a good embedding space by maximizing the number of cycle-consistent points for any pair of sequences. However that would require a differentiable version of cycle-consistency measure, two of which we introduce below.

\subsection{Cycle-back Classification}

We first compute the soft nearest neighbor $\widetilde{v}$ of $u_i$ in $V$, then figure out the nearest neighbor of $\widetilde{v}$ back in $U$. We consider each frame in the first sequence $U$ to be a separate class and our task of checking for cycle-consistency reduces to classification of the nearest neighbor correctly. The logits are calculated using the distances between $\widetilde{v}$ and any $u_k \in U$, and the ground truth label $y$ are all zeros except for the $i^{th}$ index which is set to 1.  

For the selected point $u_i$, we use the softmax function to define its soft nearest neighbor $\widetilde{v}$ as:
\begin{equation}
\widetilde{v} = \sum_j^M \alpha_j v_j, \quad where \quad \alpha_j = \frac{e^{-||u_i-v_j||^2}}{\sum_k^M e^{-||u_i-v_k||^2}}
\end{equation} 
and $\alpha$ is the the similarity distribution which signifies the proximity between $u_i$ and each $v_j \in V$. And then we solve the $N$ class (i.e.\ number of frames in $U$) classification problem where the logits are $x_k = -||\widetilde{v} - u_k||^2$ and the predicted labels are $\hat{y} = softmax(x)$. Finally we optimize the cross-entropy loss as follows:
\begin{equation}
L_{cbc} = -\sum_j^N y_j \log(\hat{y}_j)
\end{equation}

\subsection{Cycle-back Regression}

Although cycle-back classification defines a differentiable cycle-consistency loss function, it has no notion of how close or far in time the point to which we cycled back is. We want to penalize the model less if we are able to cycle back to closer neighbors as opposed to the other frames that are farther away in time. In order to incorporate temporal proximity in our loss, we introduce cycle-back regression. A visual description of the entire process is shown in Figure~\ref{fig:soft_cycle_consistency}. Similar to the previous method first we compute the soft nearest neighbor $\widetilde{v}$ of $u_i$ in $V$. Then we compute the similarity vector $\beta$ that defines the proximity between $\widetilde{v}$ and each $u_k \in U$ as:
\begin{equation}
\beta_k = \frac{e^{-||\widetilde{v}-u_k||^2}}{\sum_j^N e^{-||\widetilde{v} - u_j||^2}}
\end{equation}
Note that $\beta$ is a discrete distribution of similarities over time and we expect it to show a peaky behavior around the $i^{th}$ index in time. Therefore, we impose a Gaussian prior on $\beta$ by minimizing the normalized squared distance $\frac{|i-\mu|^2}{\sigma^2}$ as our objective. We enforce $\beta$ to be more peaky around $i$ by applying additional variance regularization. We define our final objective as:    
\begin{equation} \label{eq:1}
L_{cbr} = \frac{|i-\mu|^2}{\sigma^2} + \lambda \log(\sigma)
\end{equation}
where $\mu = \sum_{k}^N \beta_k * k$ and $\sigma^2 = \sum_{k}^N \beta_k * (k-\mu)^2$, and $\lambda$ is the regularization weight. Note that we minimize the log of variance as using just the variance is more prone to numerical instabilities. All these formulations are differentiable and can conveniently be optimized with conventional back-propagation. 

\subsection{Implementation details}
\label{sec:implementation_details}

\noindent{\bf Training Procedure}. Our self-supervised representation is learned by minimizing the cycle-consistency loss for all the pair of sequences in the training set. Given a sequence pair, their frames are embedded using the encoder network and we optimize cycle consistency losses for randomly selected frames within each sequence until convergence. We used Tensorflow~\cite{abadi2016tensorflow} for all our experiments.

\label{sec:encoding_network}
\noindent{\bf Encoding Network}.  All the frames in a given video sequence are resized to $224 \times 224$. When using ImageNet pretrained features, we use ResNet-50~\cite{he2016deep} architecture to extract features from the output of \textit{Conv4c} layer. The size of the extracted convolutional features are $14 \times 14 \times 1024$. Because of the size of the datasets, when training from scratch we use a smaller model along the lines of VGG-M~\cite{chatfield2014return}. This network takes input at the same resolution as ResNet-50 but is only 7 layers deep. The convolutional features produced by this base network are of the size $14 \times 14 \times 512$. These features are provided as input to our embedder network (presented in Table \ref{tab:small_architecture}). We stack the features of any given frame and its $k$ context frames along the dimension of time. This is followed by 3D convolutions for aggregating temporal information. We reduce the dimensionality by using 3D max-pooling followed by two fully connected layers. Finally, we use a linear projection to get a 128-dimensional embedding for each frame. More details of the architecture are presented in the supplementary material.

\begin{table}[]
\centering
\footnotesize{
\begin{tabular}{c|c|c}
\hline
Operations                  & Output Size              & Parameters\\
\hline
\hline
Temporal Stacking      & $k\times$14$\times$14$\times$$c$        & Stack $k$ context frames \\ 
3D Convolutions &  $k\times$14$\times$14$\times$512 & [3$\times$3$\times$3$, $512] $\times$ 2\\
Spatio-temporal Pooling & 512 & Global 3D Max-pooling \\
Fully-connected layers & 512& [512] $\times$ 2\\
Linear projection      & 128                        & 128                                                                                    \\ \hline
\end{tabular}
\caption{Architecture of the embedding network.}
\label{tab:small_architecture}
\vspace{-3.5em}
}
\end{table}

\section{Datasets and Evaluation}

\begin{figure*}[!t]
\centering
\includegraphics[width=\textwidth]{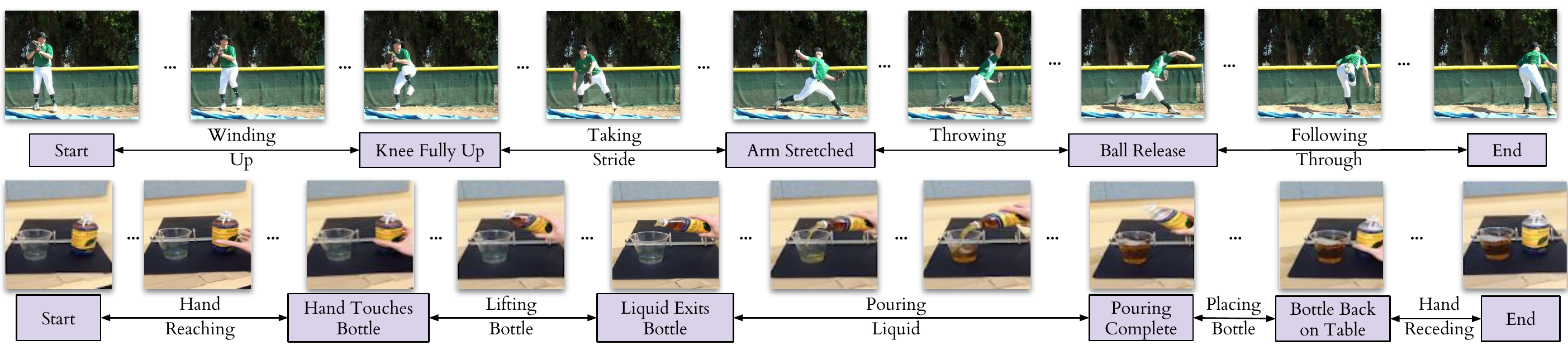}
\caption{Example labels for the actions `Baseball Pitch' (top row) and `Pouring' (bottom row). 
The key events are shown in boxes below the frame (e.g.\ `Hand touches bottle'), 
and each frame in between two key events has a phase label (e.g.\ `Lifting bottle').}
\label{fig:annotation}
\end{figure*}

We validate the usefulness of our representation learning technique on
two datasets: (i) \textit{Pouring}~\cite{Sermanet2017TCN}; and ( ii)
\textit{Penn Action}~\cite{zhang2013actemes}. These datasets both contain
videos of humans performing actions, 
and provide us with collections of videos where dense alignment can be
performed. 
While \textit{Pouring} focuses
more on the objects being interacted with, \textit{Penn Action}
focuses on humans doing sports or exercise. 

\noindent\textbf{Annotations.} For evaluation purposes, we add two types of labels to the video frames of 
these datasets: key events and phases. 
Densely labeling each frame in a video is a difficult and
time-consuming task. Labelling only {\em key events} both reduces the number of frames
that need to be annotated, and also reduces 
the ambiguity of the task (and thus the
disagreement between annotators). For example, annotators agree more
about the frame when the golf club hits the ball (a key event) than when  the
golf club is at a certain angle. The {\em phase} is the period between two key events, and all frames in the
period have the same phase label. It is similar to tasks proposed in~\cite{kuehne2014language,bojanowski2014weakly,damen2018scaling}. Examples of key events and phases are shown in 
Figure~\ref{fig:annotation}, 
and Table~\ref{tab:dataset} gives the complete list for all the actions we consider.

We use all the real videos from the \textit{Pouring}  dataset, and all but two action categories
in \textit{Penn Action}. We do not use
\textit{Strumming guitar} and \textit{Jumping rope} because
it is difficult to define unambiguous key events for these. We 
use the train/val splits of the original
datasets~\cite{Sermanet2017TCN,zhang2013actemes}. 
We will publicly release these new annotations.

\begin{table*}[!h]
\setlength{\tabcolsep}{0.3em}
\centering
\footnotesize{
    \begin{tabular}{l|c|l|c| c}
        \textbf{Action} & \textbf{Number of phases} & \textbf{List of Key Events} & \textbf{Train set size} & \textbf{Val set size} \\
        \midrule
        \midrule
        Baseball Pitch & 4 & Knee fully up, Arm fully stretched out, Ball release & 103 & 63\\
        Baseball Swing & 3 & Bat swung back fully, Bat hits ball & 113 & 57\\
        Bench-press & 2 & Bar fully down & 69 & 71\\
        Bowling & 3 & Ball swung fully back, Ball release & 134 & 85\\
        Clean and jerk & 6 & Bar at hip, Fully squatting, Standing, Begin Thrusting, Beginning Balance & 40 & 42\\
        Golf swing & 3 & Stick swung fully back, Stick hits ball & 87 & 77 \\
        Jumping jacks & 4 & Hands at shoulder (going up), Hands above head, Hands at shoulders (going down) & 56 & 56\\
        Pullups & 2 & Chin above bar & 98 & 101\\
        Pushups & 2 & Head at floor & 102 & 105\\
        Situps & 2 & Abs fully crunched & 50 & 50\\
        Squats & 4 & Hips at knees (going down), Hips at floor, Hips at knee (going up) &  114 & 116\\
        Tennis forehand & 3 & Racket swung fully back, Racket touches ball & 79 & 74\\
        Tennis serve & 4 & Ball released from hand, Racket swung fully back, Ball touches racket &  115 & 69\\
        Pouring & 5 & Hand touches bottle, Liquid starts exiting, Pouring complete, Bottle back on table & 70 & 14\\
        \bottomrule
 
    \end{tabular}
\caption{List of all key events in each dataset. Note that each action has a \textit{Start} event and \textit{End} event in addition to the key events above.}\label{tab:dataset}
}
\vspace{-1em}
\end{table*}

\subsection{Evaluation}
\label{sec:metrics}
We use three evaluation measures computed on the validation set. These metrics evaluate the model on fine-grained temporal understanding of a given action.
Note, the networks are first trained on the training set and then frozen. SVM classifiers and linear regressors are trained on the features from the networks, with no additional fine-tuning of the networks. 
For all measures a higher score implies a better model. 

\textbf{1. Phase classification accuracy:} is the per frame phase classification accuracy.
This is implemented by training a SVM classifier on  the phase labels for each frame of the
training data.

\textbf{2.\ Phase progression:}
 measures how well the \textit{progress} of a process or action is captured by the embeddings. We first define
an approximate measure of progress through a phase
as the difference in time-stamps between any
given frame and each key event. This is normalized by the number of
frames present in that video. Similar definitions can be found in recent literature \cite{ma2016learning,becattini2017done,heidarivincheh2018action}.
We use a linear regressor on the features to predict the phase progression values. It is computed as the  the average $R$-squared measure (coefficient of
determination)~\cite{wiki:rsquared}, given by:    \setlength{\belowdisplayskip}{0pt} \setlength{\belowdisplayshortskip}{0pt}\setlength{\abovedisplayskip}{1pt} \setlength{\abovedisplayshortskip}{1pt}\begin{align*}R^2 = 1- \frac{\sum_{i=1}^n (y_i - \hat{y_i})^2}{\sum_{i=1}^n (y_i - \bar{y})^2}\end{align*}where $y_i$ is the ground truth event progress value, $\bar{y}$ is the
mean of all $y_i$ and $\hat{y_i}$ is the prediction made by the linear
regression model. The maximum value of this measure is $1$.

\textbf{3.\ Kendall's Tau \cite{wiki:kendallstau}:}
is a  statistical measure that can determine how 
well-aligned two sequences are in time. Unlike the above two 
measures it does not require  additional labels for evaluation. 
Kendall's Tau is calculated over every pair of frames in a pair of videos by 
sampling  a pair of frames ($u_i, u_j$) in the first video (which has $n$
frames) and retrieving the corresponding nearest frames in the second
video, ($v_p$, $v_q$). This quadruplet of frame indices $(i,j,p,q)$ is
said to be \textit{concordant} if $i < j$ and $p < q$ or $i > j$ and
$p > q$. Otherwise it is said to be \textit{discordant}. Kendall's Tau
is defined over all pairs of frames in the first video as:
\setlength{\belowdisplayskip}{0pt}
\setlength{\belowdisplayshortskip}{1pt}
\setlength{\abovedisplayskip}{0pt}
\setlength{\abovedisplayshortskip}{1pt}
\begin{align*} \tau =
\frac{(\text{no. of concordant pairs} - \text{no. of discordant
pairs})}{\frac{n(n-1)}{2}} \end{align*}
We refer the reader to \cite{wiki:kendallstau} to check out the
complete definition. The reported metric is the average Kendall's Tau
over all pairs of videos in the validation set. It is a measure of how
well the learned representations generalize to aligning unseen
sequences if we used nearest neighbour matching for aligning a pair of
videos. A value of 1 implies the videos are perfectly aligned while a
value of -1 implies the videos are aligned in the reverse order. One
drawback of Kendall's tau is that it assumes there are no repetitive frames in a video. This might not be the case if an action is
being done slowly or if there is periodic motion. For the datasets we
consider, this drawback is not a problem.
\section{Experiments}

\subsection{Baselines}
\label{sec:methods}
We compare our representations with existing self-supervised video representation learning methods. For completeness, we briefly describe the baselines below but recommend referring to the original papers for more details. 

\noindent\textbf{Shuffle and Learn (SaL)~\cite{misra2016shuffle}.} We randomly sample triplets of frames in the manner suggested by \cite{misra2016shuffle}. We train a small classifier to predict if the frames are in order or shuffled. The labels for training this classifier are derived from the indices of the triplet we sampled. This loss encourages the representations to encode information about the order in which an action should be performed.

\noindent\textbf{Time-Constrastive Networks (TCN)~\cite{Sermanet2017TCN}.} We sample $n$ frames from the sequence and use these as anchors (as defined in the metric learning literature). For each anchor, we sample positives within a fixed time window. This gives us n-pairs of anchors and positives. We use the n-pairs loss~\cite{sohn2016improved} to learn our embedding space. For any particular pair, the n-pairs loss considers all the other pairs as negatives. This loss encourages representations to be disentangled in time while still adhering to metric constraints.

\noindent\textbf{Combined Losses.} In addition to these baselines, we can combine our cycle consistency loss with both SaL and TCN to get two more training methods: TCC+SaL and TCC+TCN. We learn the embedding by computing both losses and adding them in a weighted manner to get the total loss, based on which the gradients are calculated. The weights are selected by performing a search over 3 values $0.25, 0.5, 0.75$. All baselines share the same video encoder architecture, as described in section~\ref{sec:implementation_details}.

\subsection{Ablation of Different Cycle Consistency Losses}
We ran an experiment on the Pouring dataset to see how the different losses compare against each other. We also report metrics on the Mean Squared Error (MSE) version of the cycle-back regression loss (Equation \ref{eq:1}) which is formulated by only minimizing $|i-\mu|^2$, ignoring the variance of predictions altogether. We present the results in Table~\ref{tab:pouring_loss_ablation} and observe that the variance aware cycle-back regression loss outperforms both of the other losses in all metrics. We name this version of cycle-consistency as the final temporal cycle consistency (TCC) method, and use this version for the rest of the experiments.

\begin{table}[!h]
\setlength{\tabcolsep}{0.3em}

\centering
\footnotesize{
    \begin{tabular}{l|c|c|c}
    \toprule
      &   \textbf{Phase}  & \textbf{Phase}  & \textbf{Kendall's}\\
        \textbf{Loss} &   \textbf{Classification(\%)}  & \textbf{Progression}  & \textbf{Tau}\\
        \midrule
         Mean Squared Error & 86.16 & 0.6532 & 0.6093\\
        Cycle-back classification & 88.06 &  0.6636 & 0.6707  \\
        Cycle-back regression & \textbf{91.82} & \textbf{0.8030} & \textbf{0.8516}\\
    \bottomrule
    \end{tabular}
\caption{Ablation of different cycle consistency losses.}
\label{tab:pouring_loss_ablation}
}
\vspace{-1em}
\end{table}

\subsection{Action Phase Classification}

\noindent\textbf{Self-supervised Learning from Scratch.} We perform experiments to compare different self-supervised methods for learning visual representations from scratch. This is a challenging setting as we learn the entire encoder from scratch without labels.
We use a smaller encoder model (i.e.\ VGG-M~\cite{chatfield2014return}) as the training samples are limited.
We report the results on the \textit{Pouring} and \textit{Penn Action} datasets in Table~\ref{tab:scratch_results}. On both datasets, TCC features outperform the features learned by SaL and TCN. This might be attributed to the fact that TCC learns features across multiple videos during training itself. SaL and TCN losses operate on frames from a single video only but TCC considers frames from multiple videos while calculating the cycle-consistency loss. We can also compare these results with the supervised learning setting (first row in each section), in which we train the encoder using the labels of the phase classification task. For both datasets, TCC can be used for learning features from scratch and brings about significant performance boosts over plain supervised learning when there is limited labeled data.

\begin{table}[!h]
\centering
\renewcommand{\arraystretch}{1.1}
\begin{tabular}{p{1.25cm}|l|ccc}
\toprule
 \textbf{Datasets} & \textbf{\% of Labels $\rightarrow$} & \textbf{0.1} & \textbf{0.5} & \textbf{1.0} \\
\midrule 

{\multirow{4}{\linewidth}{\textbf{Penn Action}}} & Supervised Learning &  50.71	& 72.86 & 79.98 \\

\cline{2-5} 
& SaL~\cite{misra2016shuffle} &   66.15 &	71.10 & 72.53 \\
& TCN~\cite{Sermanet2017TCN} & 69.65 &	71.41 & 72.15\\
 & TCC (ours) & \textbf{74.68} & \textbf{76.39} & \textbf{77.30}\\
 
 \midrule

{\multirow{4}{*}{\textbf{Pouring}}} & Supervised Learning &  62.01	& 77.67 &	88.41\\

\cline{2-5} 

& SaL~\cite{misra2016shuffle} &   74.50 & 80.96 & 83.19 \\
& TCN~\cite{Sermanet2017TCN} & 76.03 & 83.27 & 84.57\\
 & TCC (ours) & \textbf{86.82} & \textbf{89.43} & \textbf{90.21}\\

\bottomrule
\end{tabular}
\footnotesize{\caption{Phase classification results when training VGG-M from scratch.}\label{tab:scratch_results}}
\vspace{-3em}
\end{table}

\begin{table}[!h]
\centering
\renewcommand{\arraystretch}{1.08}
\begin{tabular}{p{1.25cm}|l|ccc}
\toprule
 \textbf{Datasets} & \textbf{\% of Labels $\rightarrow$} & \textbf{0.1} & \textbf{0.5} & \textbf{1.0} \\
\midrule 
{\multirow{5}{\linewidth}{\textbf{Penn 
Action}}} & Supervised Learning &  67.10 & 82.78 & 86.05 \\
& Random Features & 44.18 & 46.19 & 46.81\\
& ImageNet Features & 44.96 & 50.91 & 52.86\\

\cline{2-5} 

& SaL~\cite{misra2016shuffle} & 74.87 & 78.26 & 79.96\\
& TCN~\cite{Sermanet2017TCN}  & 81.99 & 83.67 & 84.04\\
& TCC (ours) & 81.26 & 83.35 & 84.45\\
& TCC + SAL (ours) & 81.93 & 83.46 & 84.29\\
& TCC + TCN (ours) & \textbf{84.27} & \textbf{84.79} & \textbf{85.22}\\ 
 \midrule

{\multirow{5}{*}{\textbf{Pouring}}} & Supervised Learning &  75.43	& 86.14	& 91.55\\
& Random Features & 42.73 & 45.94 & 46.08\\
& ImageNet Features & 43.85	& 46.06 & 51.13\\

\cline{2-5} 

& SaL~\cite{misra2016shuffle} & 85.68 & 87.84 & 88.02\\
& TCN~\cite{Sermanet2017TCN} & 89.19 & 90.39 & 90.35\\
& TCC (ours) & \textbf{89.23} & \textbf{91.43} & \textbf{91.82}\\
& TCC + SaL (ours)& 89.21 & 90.69 & 90.75\\
& TCC + TCN (ours) &  89.17 & 91.23 & 91.51\\
\bottomrule
\end{tabular}
\caption{Phase classification results when fine-tuning ImageNet pre-trained ResNet-50.}
\label{tab:finetuning_results}
\vspace{-3em}
\end{table}

\noindent\textbf{Self-supervised Fine-tuning.} Features from networks trained for the task of image classification on the ImageNet dataset have been used for many other vision tasks. They are also useful because initializing from weights of pre-trained networks leads to faster convergence. We train all the representation learning methods mentioned in Section~\ref{sec:methods} and report the results on the \textit{Pouring} and \textit{Penn Action} datasets in Table~\ref{tab:finetuning_results}. Here the encoder model is a ResNet-50~\cite{he2016deep} pre-trained on ImageNet dataset. We observe that existing self-supervised approaches like SaL and TCN learn features useful for fine-grained video tasks. TCC features achieve competitive performance with the other methods on the \textit{Penn Action} dataset while outperforming them on the \textit{Pouring} dataset. Interestingly, the best performance is achieved by combining the cycle-consistency loss with TCN (row 8 in each section). The boost in performance when combining losses might be because training with multiples losses reduces over-fitting to cues using which the model can  minimize \textit{a} particular loss. We can also look at the first row of their respective sections to compare with supervised learning features obtained by training on the downstream task itself. We observe that the self-supervised fine-tuning gives significant performance boosts in the low-labeled data regime (columns 1 and 2).

\begin{figure}
    \centering

        \begin{subfigure}[b]{0.23\textwidth}
        \includegraphics[width=\textwidth]{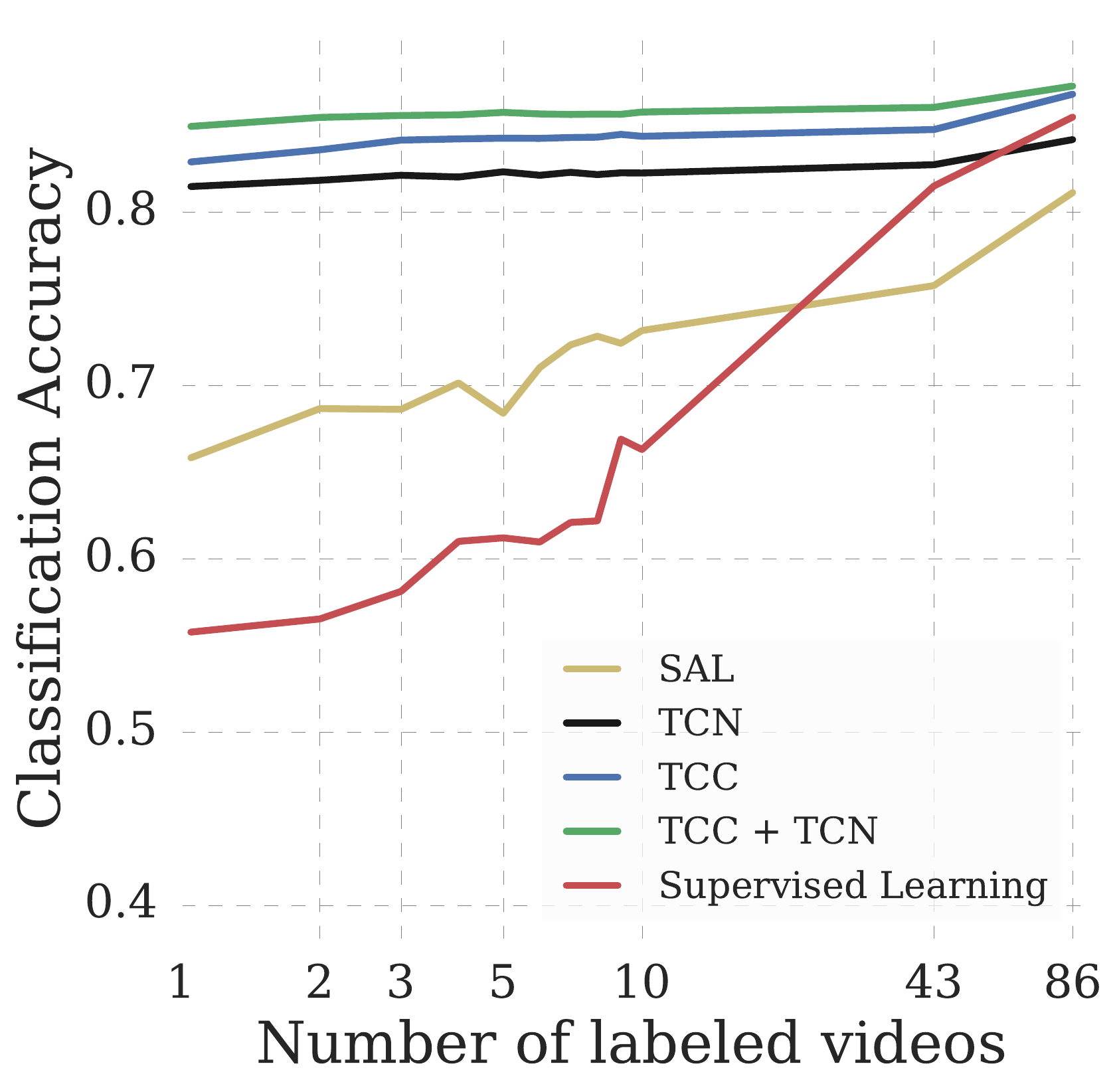}
        \caption{Golf Swing}
    \end{subfigure}
        \begin{subfigure}[b]{0.23\textwidth}
        \includegraphics[width=\textwidth]{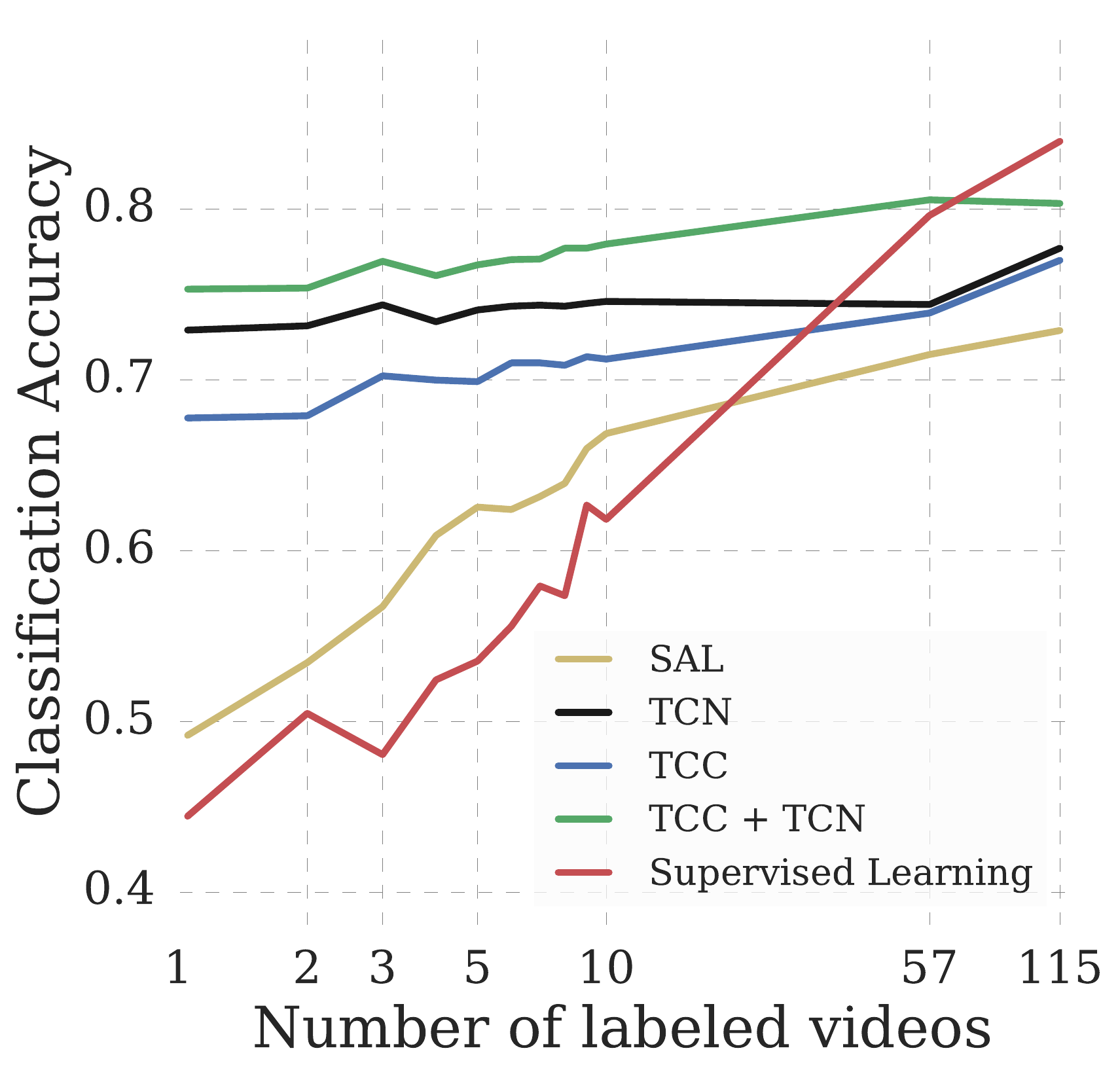}
        \caption{Tennis Serve}
    \end{subfigure}
    \caption{\textbf{Few shot action phase classification.} TCC features provide significant performance boosts when there is a dearth of labeled videos.}\label{fig:fewshot}
\end{figure}

\noindent\textbf{Self-supervised Few Shot Learning.} We also test the usefulness of our learned representations in the few-shot scenario: we have many training videos but \textit{per-frame labels} are only available for a few of them. In this experiment, we use the same set-up as the fine-tuning experiment described above. The embeddings are learned using either a self-supervised loss or vanilla supervised learning. To learn the self-supervised features, we use the entire training set of videos. We compare these features against the supervised learning baseline where we train the model on the videos for which labels are available. Note that one labeled video means hundreds of labeled frames. In particular, we want to see how the performance on the phase classification task is affected by increasing the number of labeled videos. We present the results in Figure \ref{fig:fewshot}. 
We observe significant performance boost using self-supervised methods as opposed to just using supervised learning on the labeled videos. We present results from \textit{Golf Swing} and \textit{Tennis Serve} classes above. With only one labeled video, TCC and TCC+TCN achieve the performance that supervised learning achieves with about 50 densely labeled videos. This suggests that there is a lot of untapped signal present in the raw videos which can be harvested using self-supervision.

\begin{table}[!h]

\centering
\footnotesize{

\begin{tabular}{ll|cc|cc}
\toprule
\multicolumn{2}{r|}{\textbf{Dataset $\rightarrow$}} & \multicolumn{2}{c|}{\textbf{Penn Action}} & \multicolumn{2}{c}{\textbf{Pouring}} \\
\multicolumn{2}{r|}{\textbf{Tasks $\rightarrow$}} & Progress & $\tau$ & Progress & $\tau$ \\
\midrule
\multicolumn{2}{l|}{{SL from Scratch}} &  0.5332 & 0.4997 & 0.5529	& 0.5282\\
\multicolumn{2}{l|}{{SL Fine-tuning}} &  0.6267 & 0.5582 & 0.6986 & 0.6195\\
\midrule
SaL~\cite{misra2016shuffle} & \parbox[t]{1.5mm}{\multirow{3}{*}{\rotatebox[origin=c]{90}{\scriptsize{\textbf{Scratch}}}}} &  0.4107 & 0.4940 & 0.6652 & 0.6528\\
TCN~\cite{Sermanet2017TCN} & &  0.4319 & 0.4998	& 0.6141 & 0.6647\\
TCC (ours) & & \textbf{0.5383} & \textbf{0.6024} & \textbf{0.7750} & \textbf{0.7504}\\
\midrule
SaL~\cite{misra2016shuffle} & \parbox[t]{1.5mm}{\multirow{3}{*}{\rotatebox[origin=c]{90}{\scriptsize{\textbf{Finetuning}}}}} &  0.5943 & 0.6336 & 0.7451 & 0.7331\\
TCN~\cite{Sermanet2017TCN} & &  0.6762	& 0.7328 & 0.8057 & 0.8669\\
TCC (ours) & &  0.6726 & 0.7353 & 0.8030 & 0.8516\\
TCC + SaL (ours)& & \textbf{0.6839} & 0.7286 & 0.8204 & 0.8241\\
TCC + TCN (ours) & &  0.6793 & \textbf{0.7672} & \textbf{0.8307} & \textbf{0.8779}\\
\bottomrule
\end{tabular}
}
\caption{Phase Progression and Kendall's Tau results. SL: Supervised Learning.}
\label{tab:all_regression_results}
\end{table}

\subsection{Phase Progression and Kendall's Tau}
We evaluate the encodings for the remaining tasks described in Section \ref{sec:metrics}. These tasks measure the effectiveness of representations at a more fine-grained level than phase classification. We report the results of these experiments in Table \ref{tab:all_regression_results}. We observe that when training from scratch TCC features perform better on both phase progression and Kendall's Tau for both the datasets. Additionally, we note that Kendall's Tau (which measures alignment between sequences using nearest neighbors matching) is significantly higher when we learn features using the combined losses. TCC + TCN outperforms both supervised learning and self-supervised learning methods significantly for both the datasets for fine-grained tasks.
\section{Applications}

\begin{figure}[!t]
\centering
\includegraphics[width=0.48\textwidth]{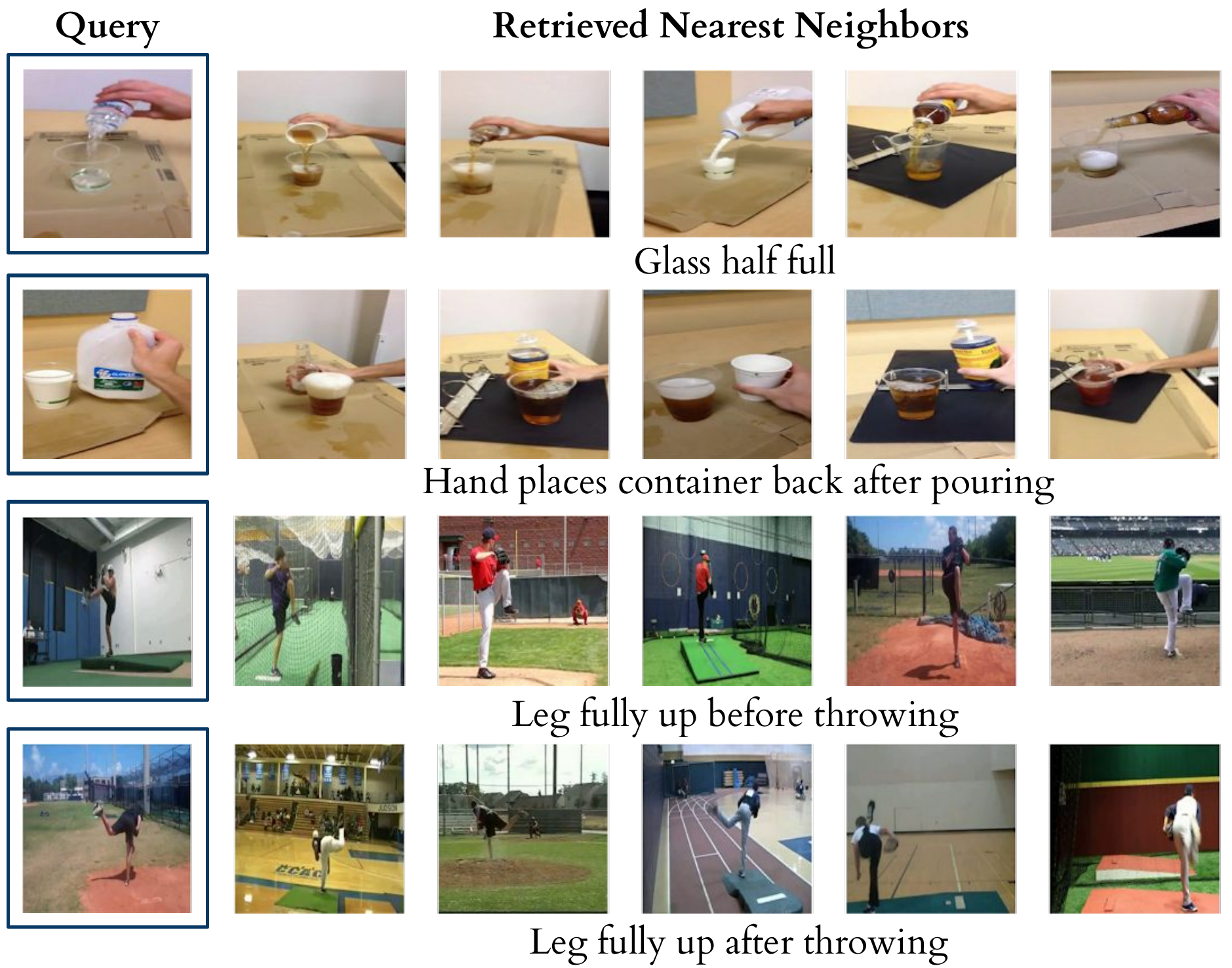}
\caption{Nearest neighbors in the embedding space can be used for fine-grained retrieval.}
\label{fig:retrieval}
\end{figure}

\begin{figure}[!t]
\centering
\includegraphics[width=0.48\textwidth]{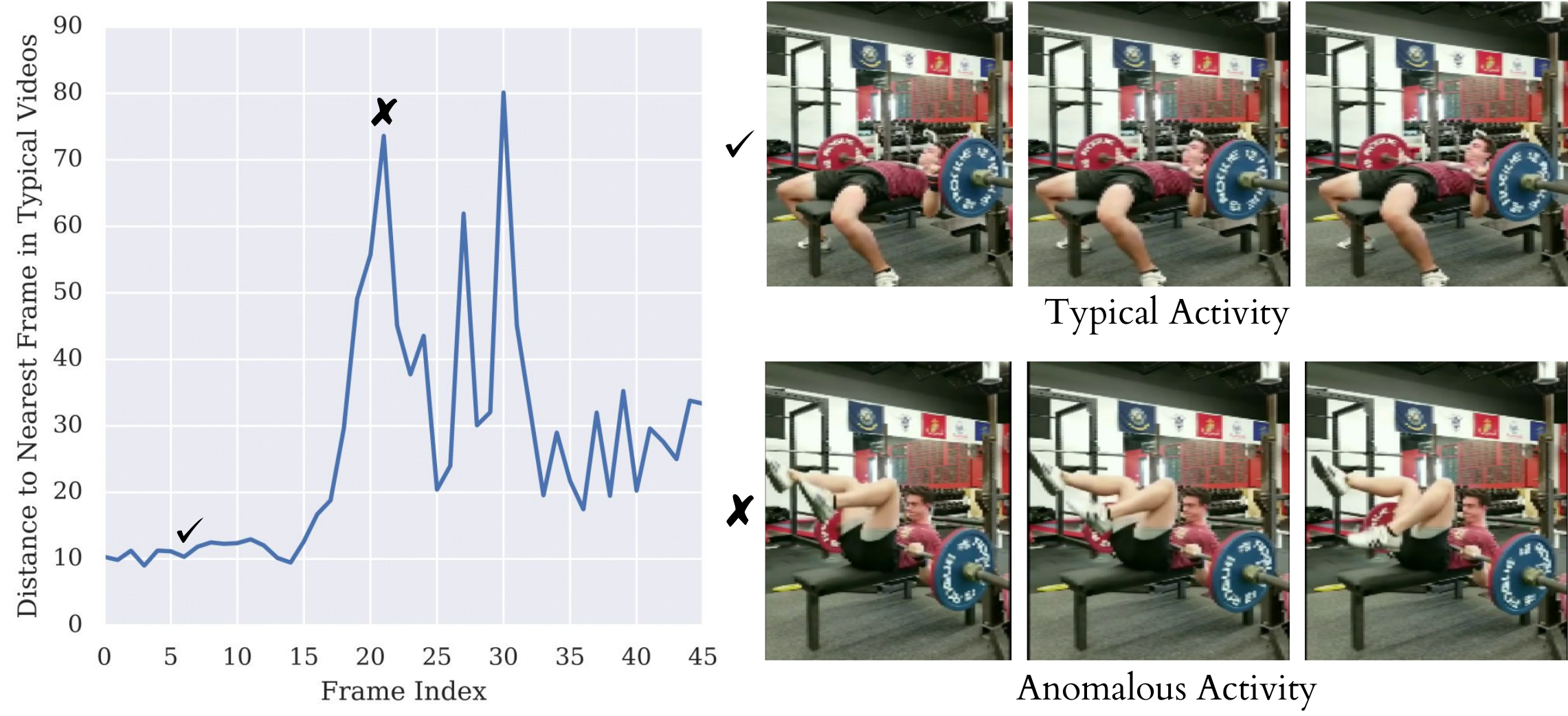}
\caption{\textbf{Example of anomaly detection in a video}. Distance from typical action trajectories spikes up during anomalous activity.}
\label{fig:anomaly}
\vspace{-2em}
\end{figure}

    \noindent\textbf{Cross-modal transfer in Videos.} We are able to align a dataset of related videos without supervision. The alignment across videos enables transfer of annotations or other modalities from one video to another. For example, we can use this technique to transfer text annotations to an entire dataset of related videos by only labeling one video. One can also transfer other modalities associated with time like sound. We can \textit{hallucinate} the sound of pouring liquids from one video to another purely on the basis of visual representations. We copy over the sound from the retrieved nearest neighbors and stitch the sounds together by simply concatenating the retrieved sounds. No other post-processing step is used. The results are in the supplementary material.

    \noindent\textbf{Fine-grained retrieval in Videos.} We can use the nearest neighbours for fine-grained retrieval in a set of videos. In Figure~\ref{fig:retrieval}, we show that we can retrieve frames when the glass is half full (Row 1) or when the hand has just placed the container back after pouring (Row 2). Note that in all retrieved examples, the liquid has already been transferred to the target container. For the \textit{Baseball Pitch} class, the learned representations can even differentiate between the frames when the leg was up before the ball was pitched (Row 3) and after the ball was pitched (Row 4). 

    \noindent\textbf{Anomaly detection.} Since we have well-behaved nearest neighbors in the TCC embedding space, we can use the distance from an \textit{ideal} trajectory in this space to detect anomalous activities in videos. If a video's trajectory in the embedding space deviates too much from the ideal trajectory, we can mark those frames as anomalous. We present an example of a video of a person attempting to bench-press in Figure \ref{fig:anomaly}. In the beginning the distance of the nearest neighbor is quite low. But as the video progresses, we observe a sudden spike in this distance (around the $20^{th}$ frame) where the person's activity is very different from the ideal bench-press trajectory.

\noindent\textbf{Synchronous Playback.} Using the learned alignments, we can transfer the pace of a video to other videos of the same action. We include examples of different videos playing synchronously in the supplementary material.
    
\section{Conclusion}

In this paper, we present a self-supervised learning approach that is able to learn features useful for temporally fine-grained tasks. In multiple experiments, we find self-supervised features lead to significant performance boosts when there is a lack of labeled data. With only one labeled video, TCC achieves similar performance to supervised learning models trained with about 50 videos. Additionally, TCC is more than a proxy task for representation learning. It serves as a general-purpose temporal alignment method that works without labels and benefits any task (like annotation transfer) which relies on the alignment itself.

\small{
\textbf{Acknowledgements}: We would like to thank Anelia Angelova, Relja Arandjelovi\'c, Sergio Guadarrama, Shefali Umrania, and Vincent Vanhoucke for their feedback on the manuscript. We are also grateful to Sourish Chaudhuri for his help with the data collection and Alexandre Passos, Allen Lavoie, Bryan Seybold, and Priya Gupta for their help with the infrastructure.
}

{\small
\bibliographystyle{ieee}
\bibliography{egbib}
}

\clearpage
\appendix

\section*{Appendix}
\section{Synchronous Playback}
One direct application of being able to align videos is that we can play multiple videos with the pace of a reference video. The task of synchronizing videos manually can be very time-consuming, often requiring multiple cuts and frame rate changes. We show how we can use self-supervised learning to reduce the effort required to synchronize videos. We present these results here: \url{https://sites.google.com/corp/view/temporal-cycle-consistency/home/visualizations-results}. We produce these videos by first embedding all frames in all the videos using our trained encoder. We choose a reference video with whose pace we want to play all the other videos. For every other video, we choose the matching frame in the whole video using dynamic time warping. This is done to enforce temporal constraints on a per-frame basis. No other post processing steps are used. 

\section{Sound Transfer}
We can also transfer other meta-data or modalities (that are synchronized with the frames in a video) only on the basis of the visual similarity. We showcase an example of such a transfer by using sound, which is arguably the most commonly available synchronized modality. Please find examples of sound transfer in the teaser video. In order to transfer the sound, we look up the nearest neighbor frames in a video that has sound. For each frame in the target video, we copy over the block of sound associated with the nearest neighbor frame. We concatenate these blocks of sound. Note, how the sound changes as the liquid flows into the container. This presents further evidence the embeddings are able to capture progress in a particular task. We use multiple frames in the sound synthesis. We average the embeddings for the multiple frames and concatenate the corresponding sounds to produce the sound blocks. We do this so that edge artifacts are reduced when we synthesize the sound for the whole video. No other post processing steps are used.

\section{t-SNE Visualization}
\begin{figure*}
    \centering
    \begin{subfigure}[b]{\textwidth}
        \includegraphics[width=\textwidth]{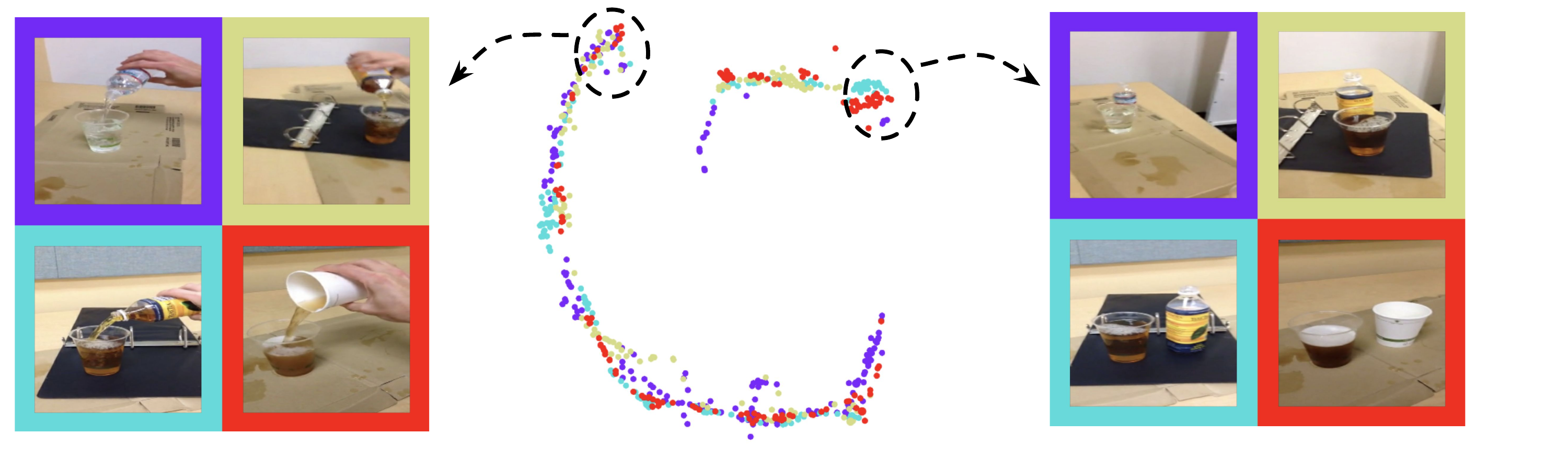}
         \caption{Pouring}
    \end{subfigure}
    \begin{subfigure}[b]{\textwidth}
        \includegraphics[width=\textwidth]{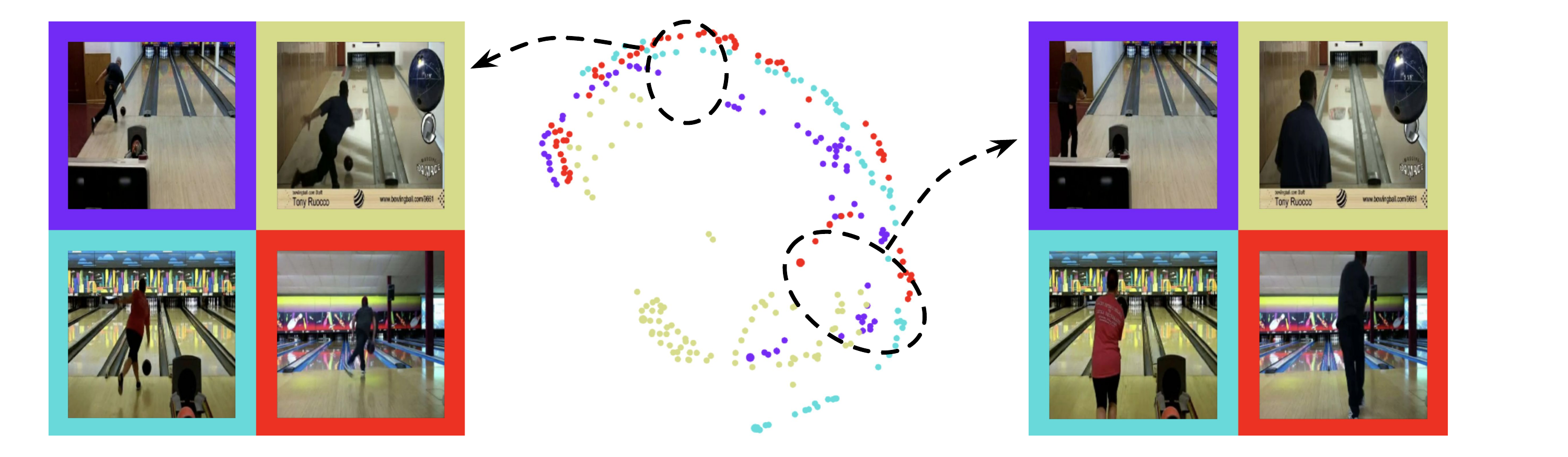}
         \caption{Bowling}

    \end{subfigure}

    \begin{subfigure}[b]{\textwidth}
        \includegraphics[width=\textwidth]{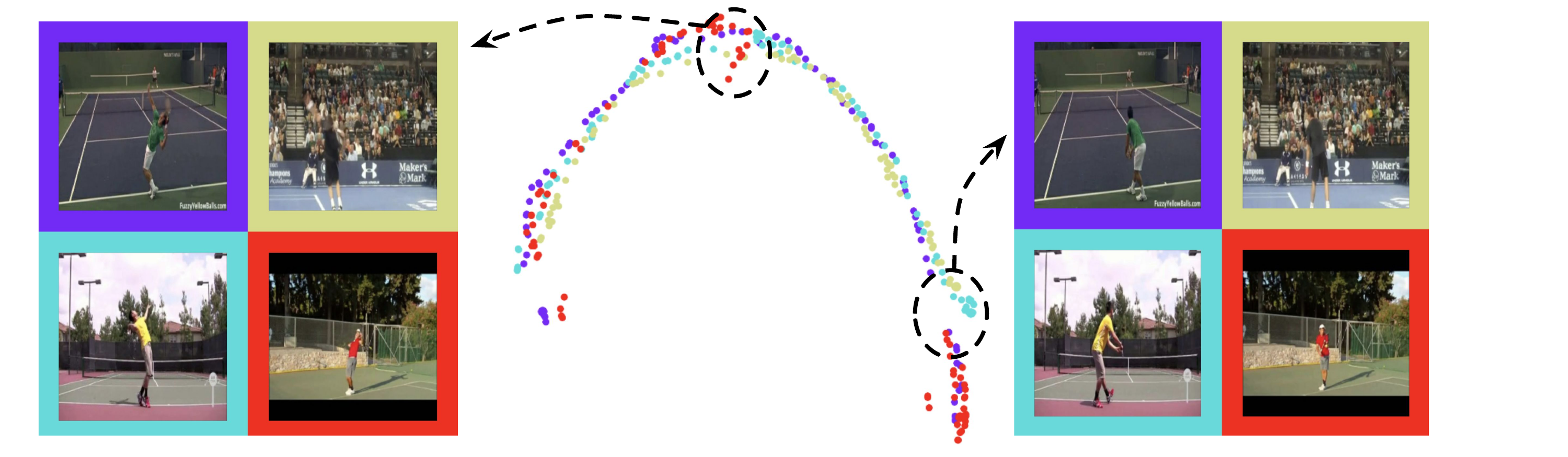}
          \caption{Tennis serve}
    \end{subfigure}
       \caption{\textbf{t-SNE Visualization of Embeddings.}}\label{fig:tsne}
\end{figure*}

We also present examples of t-SNE visualization of the embeddings in the teaser video and Figure \ref{fig:tsne}. For each action, we show trajectories of 4 videos in the embedding space. The borders are color-coded differently for each video. We sample two random time-steps for each video and show the corresponding frame and embedding location. Frames with the same border color are sampled from different time-steps in the same video. The visualization indicates how the embeddings change as an action is carried out. Additionally, they also highlight how corresponding frames from different videos in the validation set are closer to each other in the learned embedding space as compared to non-corresponding frames. This structure in the embedding space, induced by the self-supervised objectives during training, is why we are able to align different videos and perform fine-grained retrieval by simply using nearest neighbors.

\section{Fine-grained Retrieval}
\begin{figure*}
    \centering
    \includegraphics[width=\textwidth]{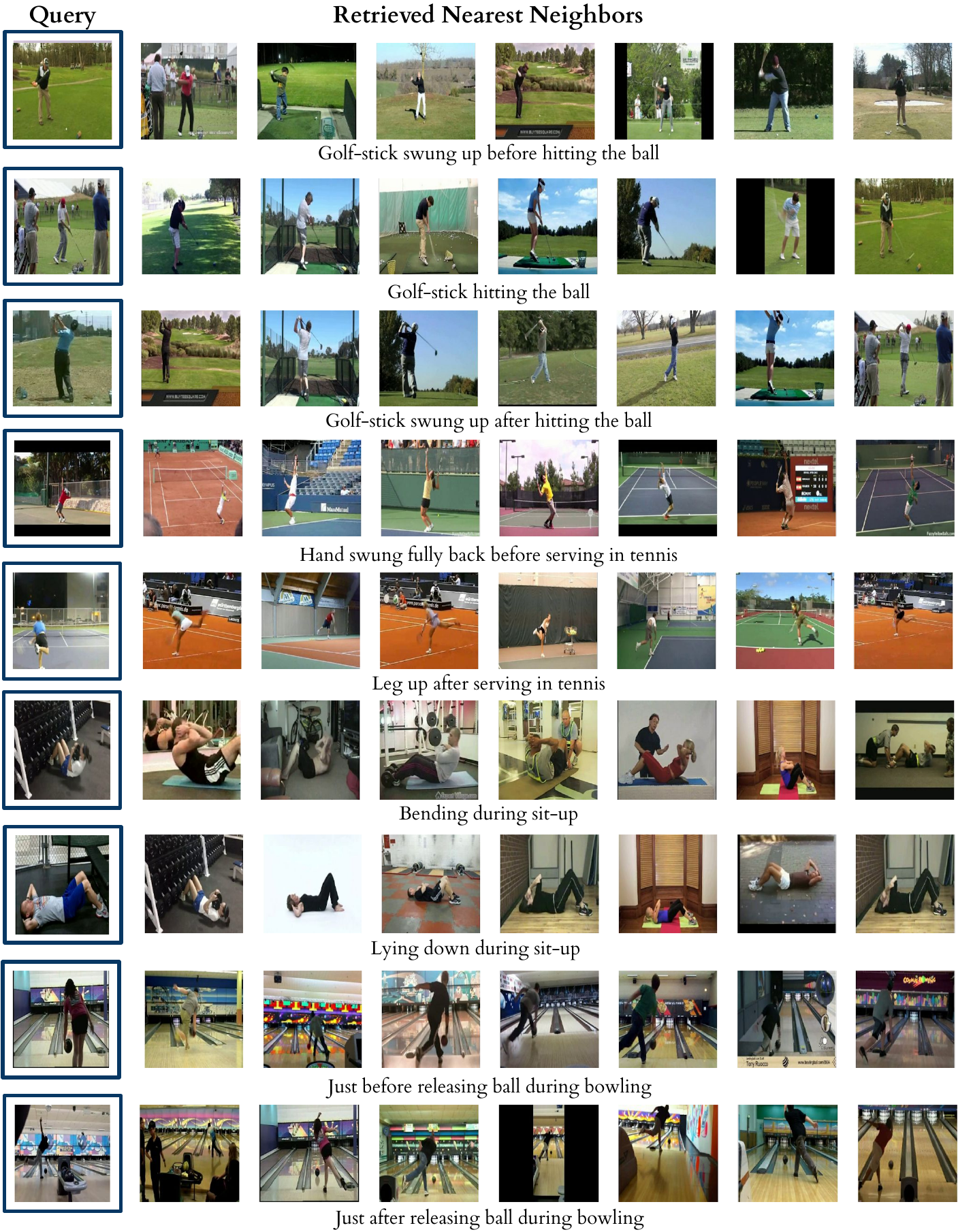}
    \caption{\textbf{Fine-grained retrieval.} Embeddings learned by temporal cycle-consistency (TCC) robustly capture fine-grained aspects of an action.}
    \label{fig:supp_finegrained}
\end{figure*}

We provide additional results for fine-grained retrieval in Figure \ref{fig:supp_finegrained}.

\section{Data Augmentation}
We use data augmentation during training. We randomly flip an entire video horizontally. We perturb brightness by adding a random number between $-32$ and $32$ to the raw pixels. We change contrast by a random factor sampled uniformly between $0.5$ and $1.5$. All training algorithms have the same data augmentation pipeline.

\section{Alignments under Different Losses}
\label{sec:similarity_matrix}
\begin{figure*}
    \centering
    \includegraphics[width=\textwidth]{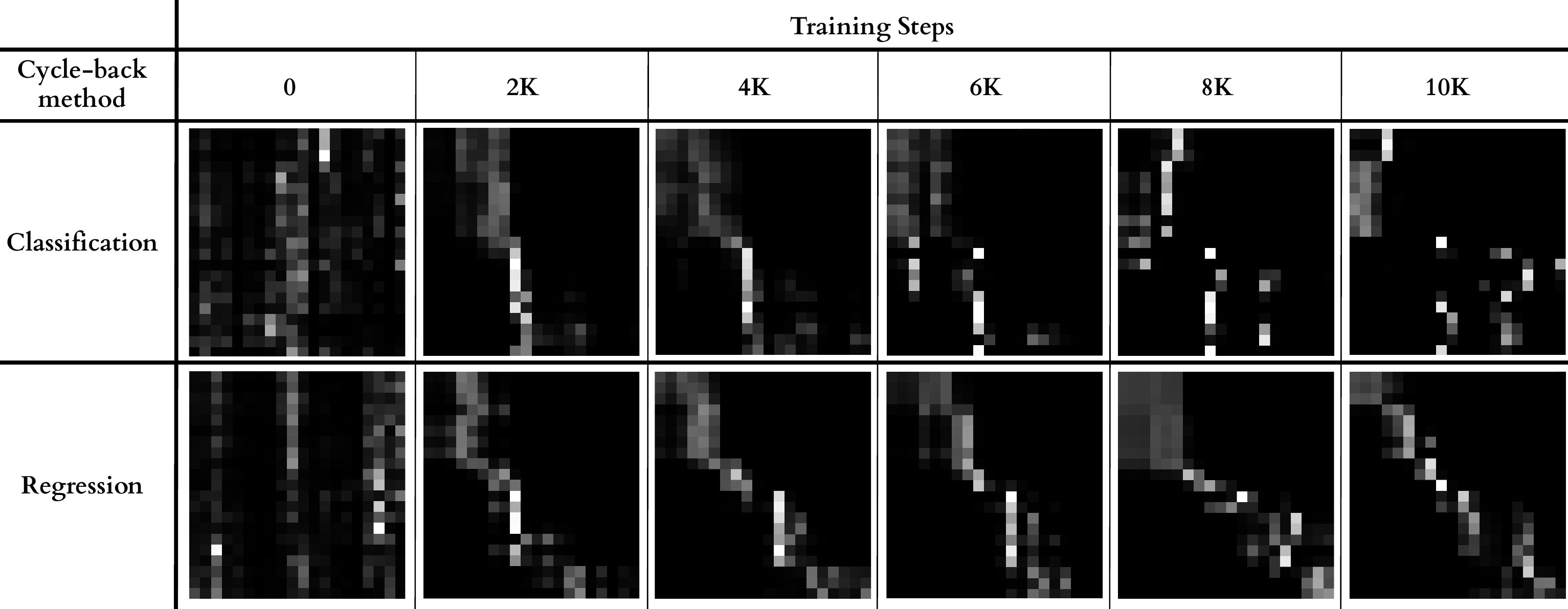}
    \caption{\textbf{Evolution of similarity matrices under different losses.} The matrices above encode the similarity  between frames of two \textit{Baseball Pitch} videos. As training proceeds, the videos get aligned more along the diagonal of the matrix, but the cycle-back regression loss is more effective at aligning the videos as compared to the cycle-back classification loss. More details in Section~\ref{sec:similarity_matrix}.}
    \label{fig:supp_similarity} 
\end{figure*}

We show how the alignment between two videos evolves as training proceeds in Figure~\ref{fig:supp_similarity}. The similarity matrices are calculated on the basis of the distance in the embedding space. The intensity at $(i,j)$ coordinates of the matrices encodes the similarity between the $i^{th}$ frame of video $1$ and $j^{th}$ frame of video $2$. The more bright a cell is, the more similar those frames are. In the beginning, the nearest neighbor matches (encoded as the brightest cells for each row/column) don't provide good alignment. As we train for more iterations, alignment between the two videos emerges as more \textit{similar} (brighter) frames exist along the diagonal. The alignment that emerges by using the cycle-back regression loss is more ordered than the cycle-back classification loss which does not take time into account while applying the cycle-consistency loss.

\section{Hyperparameters}
In Table~\ref{tab:hyperparams}, we tabulate the list of values of the hyperparameters.

\begin{table}[!h]
\setlength{\tabcolsep}{0.5em}
\renewcommand{\arraystretch}{1.15}
\centering

    \begin{tabular}{l|r}
        \toprule
        \textbf{Hyperparameter} & \textbf{Value} \\
        \midrule
        \midrule
        Batch Size & $4$ \\
        Number of frames & $20$ \\
        Optimizer & ADAM \\
        Learning Rate & $1.0 \times 10^{-4}$\\
        Weight Decay & $1.0 \times 10^{-5}$\\
        Alignment Variance $\lambda$ & $0.001$ \\
        TCN Positive Window Size & $5$\\
        Frames per second & $20$ (Penn Action), $30$ (Pouring) \\
        SaL Classifier FC Sizes & $128, 64$\\
        SaL Fraction Shuffled & $0.75$\\
        \bottomrule
    \end{tabular}
    \caption{List of hyperparameters used}
\label{tab:hyperparams}
\end{table}

\section{Architecture Details}
We describe the complete architecture of our encoder $\phi$ in Table \ref{tab:archiecture}. It is composed of 2 parts: \textit{Base Network} and \textit{Embedder Network}. The \textit{Base Network} acts on individual frames to extract convolutional features from them. Depending on the chosen base network, $c_4$ ($c$ in Table 1 of the main paper) is either $1024$ or $512$. The \textit{Embedder Network} collects convolutional features of each frame and its context window and embeds them into a single 128 dimensional vector. All the different training algorithms in our experiments are applied on top of these 128 dimensional vectors. While initially we were experimenting with larger values of $k$, we found even with $k=2$ we can get good performance on both datasets. The gap between the two frames is approximately 0.75 seconds (15 frames at 20 fps) for the Penn Action dataset and 0.3 seconds (9 frames at 30 fps) for the Pouring dataset. 

\begin{table*}[hbt!]
\centering
\renewcommand{\arraystretch}{1.50}
\begin{tabular}{c|c|c|c|c}
\hline
\textbf{Model}                             & \textbf{Layer}                  & \textbf{Output Size}                & \textbf{Pre-trained ResNet-50} & \textbf{VGG-M Like (Scratch) }                                                                            \\ \hline
\multirow{5}{*}{Base Network}     & \multirow{2}{*}{conv1} & \multirow{2}{*}{112$\times$112$\times c_1$} & \multicolumn{2}{c}{7$\times$7, 64, stride 2}                                                                  \\ \cline{4-5} 
                                  &                        &                            & \multicolumn{2}{c}{3$\times$3 max pool, stride 2}                                                                 \\ \cline{2-5} 
                                  & conv2\_x               & 56$\times$56$\times c_2$                    & \blockb{64}{256}{3}         & \begin{tabular}[c]{@{}c@{}} \blocka{128}{1} \end{tabular}    \\ \cline{2-5} 
                                  & conv3\_x               & 28$\times$28$\times c_3$                    & \blockb{128}{512}{4}            & \begin{tabular}[c]{@{}c@{}}\blocka{256}{1}\end{tabular} \\ \cline{2-5} 
                                  & conv4\_x               & 14$\times$14$\times c_4$                      & 
 \blockb{256}{1024}{3} & \blocka{512}{1}                                                                       \\ \hline
\multirow{5}{*}{Embedder Network} & Temporal Stacking      & $k  \times$14$\times$14$\times c_4$          & \multicolumn{2}{c}{Stack $k$ context frame features in time axis}                                      \\ \cline{2-5} 
                                  & conv5\_x               &  $k \times$14$\times$14$\times 512$              & \multicolumn{2}{c}{ \blockc{512}{1} }                                                                          \\ \cline{2-5} 
                                  & Spatio-temporal Pooling                        & 512                        & \multicolumn{2}{c}{Global 3D Max-Pool}                                                                            \\ \cline{2-5} 
                                  & fc6\_x                  & 512                        & \multicolumn{2}{c}{\blockd{512}{1}}                                                                               \\ \cline{2-5} 
                                  & Embedding      & 128                        & \multicolumn{2}{c}{128}                                                                                    \\ \hline
\end{tabular}
\caption{Architectures used in our experiments. The network produces an embedding for each frame (and its context window). $c_i$  depends on the choice of the base network. Inside the square brackets, the parameters in the form of: (1) $[n \times n, c]$ refers to 2D Convolution filter size and number of channels respectively (2) $[n \times n \times n, c]$ refers to 3D Convolution filter size and number of channels respectively (3) $[c]$ refers to channels in a fully-connected layer. Downsampling in ResNet-50 is done using convolutions with stride 2, while in VGG-M models we use MaxPool with stride 2 for downsampling.}
\label{tab:archiecture}
\end{table*}

\end{document}